\documentclass[leqno,11pt]{article}
\usepackage{preprint}

\usepackage{subcaption}

\usepackage{xcolor}

\usepackage{tikz}
\usetikzlibrary{arrows,positioning}
\usepackage{float}
\newcommand{\denselyConnect}[2]{
      \foreach \x in {#1}
            {\foreach \y in {#2}
                {
                \draw[->] (\x.east) -- (\y.west);
                }
            }
}

\runningauthors{Milička et al.}
\runningtitle{Semantic Space of Parts of Speech}

\title{Semantic Space of Parts of Speech}

\addauthor{1*}{Jiří Milička}{0000-0001-8605-1199}
\addauthor{2}{Ivan Kraus}{0009-0002-9686-5561}
\addauthor{1}{Arnold Stanovský}{}
\addauthor{1}{Anna Vysloužilová}{0009-0001-0670-7993}
\addauthor{1}{Barbora Štěpánková}{0000-0001-9498-7165}
\addauthor{1}{Lenka Fárová}{0000-0003-3026-2207}
\addauthor{1}{Vojtěch Cink}{0009-0000-5292-8132}
\addauthor{1}{Šárka Dohnalová}{}

 \affil{1}{Charles University, Prague, Czech Republic}

  \affil{2}{University of Potsdam, Potsdam, Germany}

\correspond{*}{Corresponding author's email: jiri@milicka.cz}

\begin{document}
\maketitle


\begin{abstract}
Parts of speech categorization is understood in the European linguistic tradition as crisp categorization, which is also reflected in corpus linguistics, where each disambiguated token is assigned exactly one POS. However, the assigned categories are largely determined by arbitrary decisions distilled into annotation manuals. Since some words stand between parts of speech in their semantics or typical syntax, and some parts of speech are closer to each other than others, POS categorization seems inherently fuzzy. We analyze this fuzziness using word2vec embeddings, training a neural network to reduce their high dimensionality to three dimensions relevant for determining parts of speech. This creates a three-dimensional space onto which we map several thousand words, revealing which are prototypical and which lie on the boundaries, and visualizing relationships between parts of speech. The study uses Universal Dependencies POS tags for French, Czech, Finnish, Russian, and English.
\end{abstract}


\begin{keywords}
parts of speech, distributional semantics, embeddings, neural network, bottleneck layer
\end{keywords}


\section{Introduction}

Parts of speech are traditionally conceived as crisp categories: while different reference works may disagree on where to classify certain words, within individual theoretical frameworks we rarely encounter situations where a specific word in a specific meaning and syntactic function is assigned to multiple categories or officially granted a ``somewhere in between'' status.

Nevertheless, we intuitively sense that some words are renegades that do not properly belong anywhere (for example, masdars in Arabic, which possess verbal semantics and valency but behave morphologically as nouns), or that certain parts of speech are closer to others (e.g., ordinal numerals in Slavic languages behave like adjectives). This fuzziness has not escaped linguists' attention, who have pointed out that classification is largely determined by tradition rather than rigorous systematicity (e.g., \cite{Haspelmath2007}). This study aims to enrich these theoretical perspectives with an empirical analysis of this fuzziness based on corpus data.

We leverage the fact that POS classification is (at least in the European tradition) based primarily on semantic and syntactic relationships, which are domains covered by embeddings \parencite{mikolov-2013-efficient}. While embeddings have been used over the past 15 years mainly for engineering-oriented natural language processing tasks, we understand them not as an ad hoc tool that happens to be convenient for our purpose, but as a consequence of Firth's and Sinclair's paradigm that builds semantic relationships upon the collocation profile of each word \parencite{firth1957synopsis,sinclair2004trust,hoey2005lexical,McEnery_Hardie_2011}.

We can conceptualize each word's collocation profile as a vector with an enormous number of dimensions (in a simplified example, each lemma that co-occurs with a given word can be understood as one dimension, with its value determined by the co-occurrence frequency). While such high-dimensional vectors are unwieldy, efficient algorithms (e.g., word2vec) can reduce them to manageable dimensionality --- typically several hundred dimensions. These dimensions do not represent anything concrete, but through the (neo-)Firthian collocability to semantics relation, they encode the semantic value of words. Consequently, vector arithmetic with these embeddings maps onto semantic arithmetic: the classic example being that subtracting the vector for \emph{man} from the vector for \emph{king} and adding the vector for \emph{woman} yields a vector very close to that of \emph{queen}. Similar semantic relations function analogously, such as \emph{Paris-France+Germany = Berlin}.

We can therefore assume that some of these hundreds of dimensions encode part of speech information, or at least provide the key to determining it. It should be possible to train a classifier on embedding --- part of speech pairs, enabling it to guess parts of speech for unknown words with reasonable accuracy. The internal representation of this classifier will constitute a function mapping semantic relationships between words onto POS categorization.

To create the classifier, we employ a neural network with a three-neuron bottleneck layer. When classifying unknown words, we examine the activation of these three neurons. This effectively reduces the several hundred embedding dimensions to just three dimensions that are relevant for determining parts of speech. We chose three dimensions as this is the maximum number that humans can easily visualize and navigate, given our lifelong experience in three-dimensional space.

These three-dimensional charts constitute the main output of this study, they neatly visualize an enormous amount of data. We attempt to describe them verbally and illustrate with selected examples in this article, but we also expect readers will explore these graphs themselves to examine aspects that interest them and draw their own conclusions.

This study is not ontological, it does not analyze what categories of words are sensible to postulate or how to define them. Rather, it examines already existing categories, particularly with regard to the Universal Dependencies framework and, more generally, the European linguistic tradition. We limit ourselves to the European linguistic tradition that derives from Latin grammatical theory because we require reasonably comparable data for analysis. We therefore selected five European languages (French, Czech, Finnish, Russian, and English). This limitation is the reason we avoid extensive typological comparison. However, if a typologist is found among the readers of this article, they are invited to further analyze the data which are fully available in a persistent OSF repository. This is also where all the scripts and tools used are to be found so that more languages can be easily analyzed.

\section{Broader context of the study} \label{sec:state-of-the-art}

The origins of part-of-speech classification date back to ancient Greece, when grammarians sought to better understand language. Even Plato distinguished between verbs and nouns, but the Greek system, which distinguishes eight parts of speech (noun, verb, participle, article, pronoun, preposition, adverb, and conjunction) and is still used today, was probably introduced by Aristarchus in the 2nd century BCE. His student, Dionysius Thrax, described this system in his grammar, 
classifying parts of speech according to a combination of their morphological, syntactic, and semantic properties. This system was later adopted by grammarians such as Apollonius Dyscolus in Greek and Priscian in Latin. (\cite{robins1966development}; \cite{nugues2006words}). Most European languages today follow this classification, with only slight modifications, distinguishing around eight parts of speech (\cite{nugues2006words}). In contrast, other linguistic traditions (e.g., Indian, Chinese, Arabic, Japanese) often distinguish a smaller number of parts of speech based on different criteria (\cite{zahnitko2019functional}).
Since the end of the 20th century, the grammatical versus semantic nature of parts of speech, and in particular the universality of parts of speech categories, has been debated (cf. \cite{Baker2003}; \cite{Croft2000}; \cite{Haspelmath2010}).

Regarding parts of speech, our study is based on the Universal Dependencies framework because we take advantage of cross-linguistic consistencies based on experience of use in many languages. Although UD follows the traditional classification of eight parts of speech, it distinguishes 17 classes of words and other elements of texts (known as UPOS --- universal part-of-speech, see Table \ref{tab:UD_POS}) and assumes that every word in any language can be assigned to one of them. 
While the definitions of these word classes vary across languages, the names remain the same, demonstrating that they have at least partially similar syntactic and, to a greater extent, semantic features \parencite{UD_zaklad}. However, the application to individual languages may follow the traditional linguistic description of the given language and bear the characteristics of different approaches\footnote{\url{https://universaldependencies.org/}} (cf. detailed information on each language studied in section \ref{sec:Results}). 

Finally, it is important to note that although no linguistic framework can function without categorization (categorization in its most general sense enables us to understand and organize the world), grammarians in the second half of the 20th century began to point out that categories often do not have neat boundaries. Word classes, therefore, do not mutually exclude one another but rather overlap. They often exhibit a clear prototypical core but a fuzzily delimited periphery, as is also evidenced by the data presented in this study (\cite{aarts2006conceptions}).

\begin{table}[H]
    \centering
    \begin{tabular}{l l l l}
         \textbf{Traditional POS} & \textbf{UPOS} & \textbf{Category}
         \\\midrule
         noun & NOUN & common noun\\ 
         & PROPN & proper noun\\ 
         verb & VERB & main verb\\
&AUX & auxiliary verb or other tense, aspect, or mood particle\\
adjective & ADJ & adjective\\
& DET & determiner (including article)\\
& NUM & numeral (cardinal)\\
adverb & ADV & adverb\\
pronoun & PRON & pronoun\\
preposition & ADP & adposition (preposition/postposition)\\
conjunction & CCONJ & coordinating conjunction\\
& SCONJ & subordinating conjunction\\
interjection & INTJ & interjection\\
--- & PART & particle (special single word markers in some languages)\\
--- & X & other (e.g., words in foreign language expressions)\\
--- & SYM & non-punctuation symbol (e.g., a hash (\#) or emoji)\\
--- & PUNCT & punctuation\\
    \end{tabular}
    \caption{Universal POS}
    \label{tab:UD_POS}
\end{table}

\section{Data} \label{sec:data}



We use embeddings provided by the NLPL word embeddings repository\footnote{\url{https://vectors.nlpl.eu/repository/}} \parencite{fares-etal-2017-word}.  We opted for embeddings trained using the word2vec Continuous Skipgram algorithm \parencite{mikolov-2013-efficient} with the context window of size 10. Since word2vec embeddings are static, they allow us to look at word forms without the need for us to provide any additional texts. All the embeddings we used are 100-dimensional and trained on unlemmatized CoNLL17 corpora \parencite{conll-2017-shared-task}, consisting of texts obtained from CommonCrawl and Wikipedia,\footnote{\url{https://universaldependencies.org/conll17/data.html}} which also makes it easier to draw comparisons between embeddings for various languages.



In order to reduce the overall data size, we selected top 10\,000 most frequent forms for each language. We used the InterCorp v16ud \parencite{cermak-rosen-2012}, a multilingual parallel corpus with automatic UD annotation, to compile form-frequency dictionaries for each language we analyze. We excluded certain UPOS from our analyses, because they denote non-word tokens (PUNCT, SYM) or because they tend to be assigned to strange-looking tokens, often misclassifying them (INTJ, PROPN, X). Since each token in the corpus is already assigned a POS label, we were able to exclude the undesirable UPOS from our frequency dictionaries already in the compilation phase. 

We then used the dictionary to create vectors describing POS each form occurs as. Each of the vectors is made of 10--12 items, depending on the language, for all the UPOS we ultimately consider.\footnote{NOUN, VERB, AUX, ADJ, DET, NUM, ADV, PRON, ADP, CCONJ, SCONJ, PART. In InterCorp, some of these categories remain unused for some languages.} The items are numbers between 0 and 1 (inclusive), denoting the relative frequency with which the given form occurs as the given UPOS in the filtered InterCorp corpus. The items add up to 1 for each form as a result. 

In the charts below, we color word forms after the POS the forms were most frequent as. 

\section{Methodology} \label{sec:methodology}

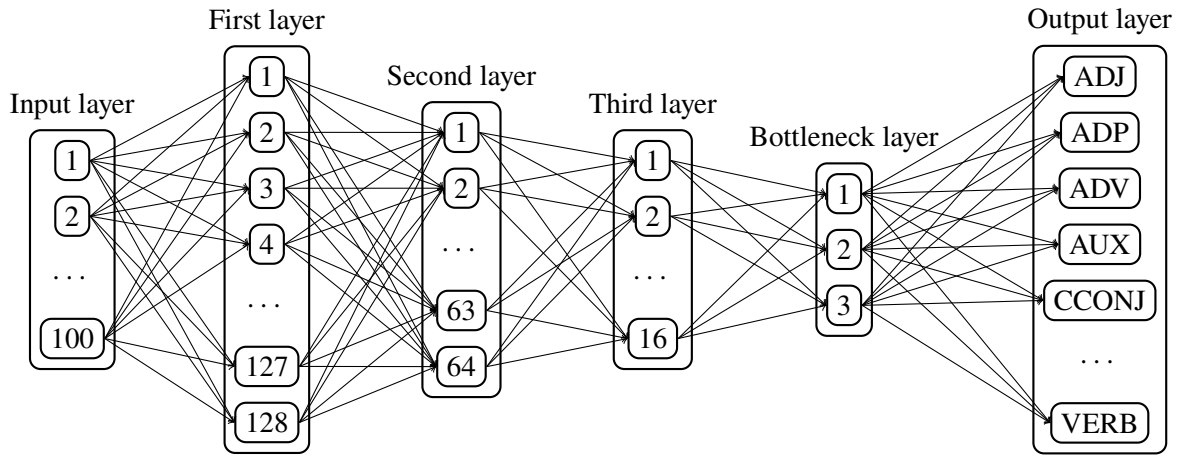
\begin{figure}[H]
    \centering
    \begin{tikzpicture}[
        node distance=2cm,    
        bigbox/.style={draw, thick, rounded corners, rectangle, row sep=0.5em, anchor=center}
        ]
        \matrix[label=above:Input layer, bigbox] (input-layer) {
            \node[rectangle, draw] (0101){1};\\
            \node[rectangle, draw] (0102) {2};\\
            {}\\
            \node {\ldots};\\
            {}\\
            \node[rectangle, draw] (0199) {100};\\
        };
        
        \matrix [label=above:First layer,right=of input-layer, bigbox ] (first-layer){
            \node[rectangle, draw] (0201) {1};\\
            \node[rectangle, draw] (0202) {2};\\
            \node[rectangle, draw] (0203) {3};\\
            \node[rectangle, draw] (0204) {4};\\
            {}\\
            \node {\ldots};\\
            {}\\            
            \node[rectangle, draw] (0298) {127};\\
            \node[rectangle, draw] (0299) {128};\\
        }; 
        \matrix [label=above:Second layer, right=of first-layer, bigbox] (second-layer){
            \node[rectangle, draw] (0301) {1};\\
            \node[rectangle, draw] (0302) {2};\\
            {}\\
            \node {\ldots};\\
            {}\\            
            \node[rectangle, draw] (0398) {63};\\
            \node[rectangle, draw] (0399) {64};\\        
        };
        
        \matrix [label=above:Third layer, right=of second-layer, bigbox] (third-layer){
            \node[rectangle, draw] (03501) {1};\\
            \node[rectangle, draw] (03502) {2};\\
            {}\\
            \node {\ldots};\\
            {}\\            
            \node[rectangle, draw] (03599) {16};\\      
        };
        
        \matrix [label=above:Bottleneck layer, right=of third-layer, bigbox] (bottleneck-layer){
            \node[rectangle, draw] (0401) {1};\\
            \node[rectangle, draw] (0402) {2};\\
            \node[rectangle, draw] (0403) {3};\\
        }; 
        \matrix [label=above:Output layer, right=3cm of bottleneck-layer, bigbox] (output-layer){
            \node[rectangle,draw] (0501){ADJ};\\
            \node[rectangle,draw] (0502){ADP};\\
            \node[rectangle,draw] (0503){ADV};\\
            \node[rectangle,draw] (0504){AUX};\\
            \node[rectangle,draw] (0505){CCONJ};\\
            {}\\
            \node {\ldots};\\
            {}\\  
            \node[rectangle,draw] (0512){VERB};\\
        }; 
        \denselyConnect{0101, 0102, 0199} {0201, 0202, 0203, 0204, 0298, 0299} 
        \denselyConnect{0201, 0202, 0203, 0204, 0298, 0299} {0301, 0302, 0398, 0399}
        \denselyConnect{0301, 0302, 0398, 0399} {03501, 03502, 03599}
        \denselyConnect{03501, 03502, 03599} {0401, 0402, 0403}
        \denselyConnect{0401, 0402, 0403} {0501, 0502, 0503, 0504, 0505, 0512}
    \end{tikzpicture}
    \caption{Layout of the neural network used in the experiment}
    \label{fig:nn-layout}
\end{figure}

Having acquired POS annotations for the selected word forms in vector form, as well as their embeddings, we proceeded to prepare the data for the training of our models. First, we split the data in two sets ($A$ and $B$) of roughly the same size, ensuring that all forms of a given lemma are only present in one set.\footnote{We assume that different forms of the same lemma might be very similar to each other and thus potentially introduce the risk of training data leakage into the validation dataset.} This allowed us to use one as training data and the other as validation data, training two neural networks, $NN_A$ (trained on $A$, validated against $B$) and $NN_B$ (trained on $B$, validated against $A$). The networks receives the embeddings and classifies them in terms of UD POS.

$NN_A$ and $NN_B$ technically differ in training data used, their structure is the same. However, due to the two sets of training data containing significantly different shares of mainly closed-class POS, it may happen (often happens, in fact) that some POS end up not being represented in the network at all. Both networks use the embeddings as input, passing them through the 128-neuron first hidden layer, 64-neuron second hidden layer, and 16-neuron third hidden layer. To help us visualize the relationships between different parts of speech, we add a 3-neuron bottleneck layer before the final classification layer. The three hidden layers are ReLU activated, with dropout rate of 0.3. Each neural network was trained for 30 epochs.

To visualize the collected data, we take the activation values of the bottleneck layer for each wordform and map them onto a 3D space. 
We perform no further transformations.

Both the source code and the resulting data are available at \url{https://osf.io/gm6zd/}.

\section{Results}
\label{sec:Results}

We provide brief analyses of the resulting projections for each of the 5 languages, together with commentaries on their POS traditions. To provide the reader with basic reference, we also include charts (2 per set, meaning 4 per language in total) of the visualizations taken from manually selected camera angles. However, we encourage the reader to explore the interactive visualizations by themselves, since the 2D representation does not capture the real shape and individual words are represented just by a dot so that the word label cannot be seen. The full 3D interactive charts can be viewed on \url{https://hypertext.cz/pos/} (large file) and also obtained from the Support Information and from OSF repository, together with respective confusion matrices and the source code.

\definecolor{colorADJ}{HTML}{e60049}
\definecolor{colorADP}{HTML}{dc0ab4}
\definecolor{colorADV}{HTML}{50e991}
\definecolor{colorAUX}{HTML}{fd7f6f}
\definecolor{colorCCONJ}{HTML}{9b19f5}
\definecolor{colorDET}{rgb}{1, 0.639, 0}
\definecolor{colorNOUN}{rgb}{0.043, 0.706, 1}
\definecolor{colorNUM}{rgb}{0.702, 0.831, 1} 
\definecolor{colorPART}{HTML}{00bfa0}
\definecolor{colorPRON}{HTML}{77ac35}
\definecolor{colorSCONJ}{HTML}{7c1158}
\definecolor{colorVERB}{HTML}{e6d800}
\newcommand{\legendcolorbox}[1]{\fcolorbox{#1}{#1}{\phantom{I}}}
\newcommand{\legend}{
    \legendcolorbox{colorNOUN} NOUN
    \legendcolorbox{colorVERB} VERB
    \legendcolorbox{colorAUX} AUX
    \legendcolorbox{colorADJ} ADJ
    \legendcolorbox{colorDET} DET
    \legendcolorbox{colorNUM} NUM
    \legendcolorbox{colorADV} ADV
    \legendcolorbox{colorPRON} PRON
    \legendcolorbox{colorADP} ADP
    \legendcolorbox{colorCCONJ} CCONJ
    \legendcolorbox{colorPART} PART
    \legendcolorbox{colorSCONJ} SCONJ
}
\newcommand{\legendnoPART}{
    \legendcolorbox{colorNOUN} NOUN
    \legendcolorbox{colorVERB} VERB
    \legendcolorbox{colorAUX} AUX
    \legendcolorbox{colorADJ} ADJ
    \legendcolorbox{colorDET} DET
    \legendcolorbox{colorNUM} NUM
    \legendcolorbox{colorADV} ADV
    \legendcolorbox{colorPRON} PRON
    \legendcolorbox{colorADP} ADP
    \legendcolorbox{colorCCONJ} CCONJ
    \legendcolorbox{colorSCONJ} SCONJ
}
\newcommand{\legendnoPARTDET}{
    \legendcolorbox{colorNOUN} NOUN
    \legendcolorbox{colorVERB} VERB
    \legendcolorbox{colorAUX} AUX
    \legendcolorbox{colorADJ} ADJ
    \legendcolorbox{colorNUM} NUM
    \legendcolorbox{colorADV} ADV
    \legendcolorbox{colorPRON} PRON
    \legendcolorbox{colorADP} ADP
    \legendcolorbox{colorCCONJ} CCONJ
    \legendcolorbox{colorSCONJ} SCONJ
}

\newcommand{\chartQ}[3]{
\begin{figure}[h!]
\centering
    \begin{subfigure}{\textwidth}
    \includegraphics[width=0.45\textwidth]{images/#1/#111.pdf}\hfill
    \includegraphics[width=0.45\textwidth]{images/#1/#112.pdf}
    \caption{Set A}
    \end{subfigure}
    
    \vspace{1em}
    
    \begin{subfigure}{\textwidth}
    \includegraphics[width=0.45\textwidth]{images/#1/#121.pdf}\hfill
    \includegraphics[width=0.45\textwidth]{images/#1/#122.pdf}
    \caption{Set B}
    \end{subfigure}
    
\caption*{#3}
\caption{The mapping of parts of #2 speech --- bottleneck layer}
\label{Chart:NNQ_#1}
\end{figure}
}

\subsection{French}

\subsubsection{French POS}
Definitions of POS in French are generally based on morphology, semantics and syntax. French grammars commonly define nine parts of speech:  \textit{nouns}, \textit{adjectives}, \textit{articles}, \textit{pronouns}, \textit{verbs}, \textit{adverbs}, \textit{prepositions}, \textit{conjunctions} and \textit{interjections}.  There is some dispute about the definitions, some grammars also define a separate POS category of \textit{presentatives/introducers}\footnote{It needs to be noted that presentatives and introducers are not synonyms, though they are labels for the same POS, their definitions slightly vary. (\cite{grevisse-2007}, \cite{grammairemethodique})}, a closed category of words that present a word or a phrase (e.g. \textit{\textbf{Voici} votre manteau} `\textbf{Here's} your coat') Some grammars also define coordinating and subordinating conjunctions as separate POS.

\subsubsection{French UD}
French UD classification varies in numerous ways from French grammar approaches: 
Articles are divided into the DET and PRON categories. French UD also does not feature a tag for presentatives/introducers, which are tagged as VERBs or PREPs, phrasal presentatives (e.g. \textit{Il y a}, `there is/there are'\footnote{The most common translation equivalents are used in the text. We note that particular lexical units could be translated differently in different contexts.}) are delimited. UD also defines three auxiliary verbs: \textit{être} 'to be', \textit{avoir} 'to have', and \textit{faire} `to make'. \textit{Faire} is used in causative constructions but is generally not defined as an auxiliary verb in French grammars. Conjunctions are divided into CCONJ and SCONJ, which is in accordance with some French grammars, while other define them as a single POS. Cardinal numerals are defined as a separate POS category NUM in UD, while French grammar defines them as nouns, numeral determiners or as numeral adjectives (\cite{grevisse-2007}), based on usage. Other numerals, such as ordinal numerals, are tagged as ADJs or ADVs. The PART category is not used in the French InterCorp annotation, hence its absence. It is also to be noted that French UD tokenization does delimit tokens connected by a hyphen, except in cases where they form a single word (e.g. \textit{États-Unis}, `the United States'). However, some expressions with a hyphen that should be separated are not (e.g. \textit{c'est-à-dire}, `that is to say', formed by four different words), and sometimes tend to be misclassified. There are also some contractions that are considered multi-word tokens and are segmented into individual syntactic words (e.g. \textit{duquel}, a contraction of the words \textit{de}, \textit{le} and \textit{quel}, is divided into three tokens).

\chartQ{FR}{French}{\legendnoPART}

\subsubsection{Analysis}
The visualisation of the French data shows clear distinction between VERBs, ADJs, NOUNs and NUMs, with a less clear, small tentacle\footnote{Since all POS charts presented here are octopus-shaped, we developed octopus-related nomenclature during internal discussions.} of ADVs. Some PRONs\footnote{Imperative verb forms are sometimes formed with a disjunctive pronoun connected by a hyphen (e.g. \textit{dis-moi}, `tell me`), which, as mentioned above, UD sometimes does not delimit, thus causing it to misclassify these words as pronouns in the visualisations.} form a group within a cluster of SCONJs, CCONJs and DETs, other are blended within the ADJ tentacle. 

\textbf{Transitions/blending between tentacles}: Blending occurs between all three main tentacles of VERBs, ADJs and NOUNs. The VERB and ADJ tentacles blend because of polyfunctional word forms. In French, many word forms do not correspond to a single POS when not in the context of a sentence. For example, the word \textit{oublié} can be considered as ADJ (`forgotten') or be part of a compound past form formed with an auxiliary verb (\textit{j'ai oublié}, `I forgot', from \textit{oublier}). ADJ and NOUN tentacles blend mainly because of homonymy and polyfunctionality of certain words (e.g. \textit{inconnu}, `unknown', used as both an adjective and a noun describing an unknown person'). VERB and NOUN tentacles also blend due to homonymy (e.g. \textit{pêcher}, meaning `peach tree' but also the infinitive form of `to fish'), or polyfunctionality of the word itself. 
Also, even though ADVs seem to form their own grouping, some adverbs blend with the ADJ tentacle due to their frequent, near-exclusive collocation with adjectives (e.g. \textit{purement}, `purely', \textit{exclusivement}, `exclusively'). Possessive and interrogative DETs and possessive PRONs are completely blended within the ADJ tentacle. Interestingly, plural forms of possessive PRONs are more blended into the tentacle than singular forms.

\subsubsection{French conclusion}
Overall, the three main tentacles of NOUNs, VERBs and ADJs are the most prominent, which is not surprising. The blending of these three tentacles is motivated mostly by polyfunctionality and homonymy. The fact that numerals do form a separate group within the CONJs, ADPs and PRONs cluster is quite surprising, given their definition of determiners or adjectives. While they do act as such in the context of a sentence, it's clear that semantically, they behave differently. 

\subsection{Czech}

\subsubsection{Czech POS}
While the traditional Czech approach to parts of speech was based on a combination of morphology, syntax and semantics, in more recent grammars (e.g. \cite{mluvniceII}) it has been replaced by the relation between semantics and functionality (i.e. syntax). Based on their primary/prototypical combination, four\textit{ basic} parts of speech are defined: \textit{nouns} (as subject or object), \textit{verbs} (as predicate), \textit{adjectives} (as declension attribute) and \textit{adverbs} (as various circumstantial relations, including space, time, manner). All other combinations (e.g. an adverb or a noun as part of a predicate) are potentially borderline/problematic. Similarly, other inflected words, i.e. \textit{numerals} and \textit{pronouns}, can function as nouns, adjectives or adverbs, but still have their own POS. 
Furthermore, there are three other so-called \textit{functional} POSs: \textit{conjunctions}, \textit{prepositions}, and \textit{particles},\footnote{In Anglo-American approaches it corresponds to the term pragmatic/adverbial marker.} and a special POS for \textit{interjections}.

\subsubsection{Czech UD}
In the Czech version of UD,\footnote{\url{https://universaldependencies.org/cs/}} in comparison with the Czech grammar approach, there are the following peculiarities:
As for \textit{numerals} and \textit{pronouns}, in UD more attention is paid to their syntactic and morphological behaviour. Words traditionally called \textit{pronouns} are divided into two categories: substantive pronouns (UD tag PRON) and attributive pronouns, including pronominal numerals (UD tag DET).\footnote{Czech does not have articles, and Czech grammar does not use a term determiners.} Cardinal numerals have their own tag NUM, other types are tagged ADJ or ADV, based on their syntactic and morphological behavior. Czech \textit{conjunctions} are divided into two POSs: CCONJ and SCONJ. Finally, there is also a special POS AUX for \textit{auxiliary verbs}, in Czech UD, the verb \textit{být} `to be' (and its variant \textit{bývat} `to usually be') is considered the only representative of this category.

\chartQ{CS}{Czech}{\legend}

\subsubsection{Analysis}
The visualisation of the Czech data shows clear tentacles of VERBs, NOUNs, ADJs and NUMs, another tentacle, ADV, is smaller but still visible. At the confluence of the NOUN and ADJ tentacles, PRON and DET form a separate island/group, DET is closer to ADJ (as expected), but the split is not pronounced.
\textbf{Transitions/blending between tentacles}:
 The transition/blending between ADJ and NOUN occurs mainly in the form of typical collocations (e.g. \textit{krevní} `blood', \textit{dopravy}, `transport' gen. sg.), sometimes mono-colocability (e.g. \textit{pozměňovací} `amendatory'), and in the form of ellipses, where the original adjective also begins to behave as a NOUN (e.g. \textit{vedoucí} `leading', \textit{kuřecí} `chicken', \textit{mrtví} `dead'). It can also be a case of formal homonymy (e.g. \textit{povýšení} `promotion' noun neutr., and `haughty' adj. masc.). The transition/blending between ADJ and VERB can be seen mainly in the short forms 
 of adjectives and passive participles (the difference is often one sound or the quantity of a vowel, both often occur with the verb \textit{být} `to be', their meaning can sometimes be the same (e.g. \textit{spokojená} `satisfied', \textit{přesvědčena} `convinced')). 
 The peculiarity of the space between VERB and NOUN is that it contains mainly parts of phrasal verbs (e.g. \textit{dávat přednost} `to give preference') and then words that occur in separate statements, e.g. addressing in the vocative case (e.g. \textit{chlapče} `boy', \textit{maminko} `mother'), exclamation, command (imperative) (e.g. \textit{sklapni} `shut up').
As an interesting fact --- the various forms of the verb \textit{být} `to be' (AUX) run along the whole length of the VERB tentacle --- it may be related to the polyfunctionality of the verb --- it not only makes auxiliary forms (future, past, etc.) but also the copula, and thus gets into a relationship with both verbs and nominals.
The middle cluster/clump consists mainly of function/grammatical words. Conjunctions, particles and secondary prepositions are often polyfunctional/homonymous. Perhaps only some primary prepositions are closer to NOUNs or VERBs due to their frequent co-occurrence.

\subsubsection{Czech conclusion}
Not surprisingly, the VERB, ADJ and NOUN tentacles are prominent, which in Czech is supported by their syntactic, morphological and semantic behaviour. Transitions/blending between them correspond to homonymy, word form transitions or use within idioms, phrases. The ADV tentacle shows that a base of this POS is formed by meaningful (non-deictic) words, which can usually be graded and negated. A separate tentacle for NUM is somewhat surprising. Its central part consists of cardinal numerals written in words in various cases, and towards the centre of the cluster/clump there are numerals that also function as pronouns or names (e.g. \textit{jeden }`one', \textit{oba} `both', \textit{půl} `half'). This confirms that NUM should be a separate POS (which is sometimes doubted in Czech). The situation with functional and grammatical words is less clear --- in addition to their limited semantics, this may be due to their limited number in the language.

\subsection{Finnish}

\subsubsection{Finnish POS}
The Finnish POS are morphologically and syntactically based, but they are also based on at least some semantic similarity. The discussion on the borders between Finnish POS has been going on since the beginning of the 20th century and is certainly not over (\cite{pentti}; \cite{haku}). The main criterion of POS in Finnish is usually inflection, which divides Finnish words into three main categories: nominal with inflection in number and case, verbs with person, tense and modal inflection and particles with no or only partial inflection. Inflection distinguishes between nominals and verbs, although verb infinitives have partial case inflection and participles have full number and case inflection (\cite{iso}). Due to large lexicalization boundaries between the POS categories are quite blurred (e.g. \textit{loukkaantunut} `injured' can be used as a part of a verb form, as a noun or as an adjective).

The division into subcategories within nominals is based on syntactic and semantic grounds: nouns have full inflection, including possessive suffixed forms, one of the main functions of adjectives is to act as congruent attributes of nouns, pronouns which have low descriptive content and can occur in syntactic positions specific to both nouns and adjectives, and numerals create their own category based on semantics (\cite{iso}). There is though an overlap between some pronouns and numeral, as well as some nouns and adjectives, both in form and function.

Similarly, there are no clear borders between some adverbs and adpositions, conjunctions and adverbs, there are also words which are difficult to place in any category of POS, such as \textit{itse} `self', which has properties of a pronoun, adverb, particle and even a noun in different uses.

There are also features that unite members of otherwise distinct word classes. This is the case of gradation, which apply not only to a large number of adjectives but also to many adverbs and some words that are also counted as pronouns (e.g. \textit{useammin} `more often'). Some pronouns and some adverbs, on the other hand, share the ability to express relative quantity, e.g. (\textit{ei}) \textit{ketään} `nobody', and \textit{monta} (`many' --- PAR sg. from \textit{moni}) and \textit{paljon} `many/much'.

To sum up: the same word can often belong to more than one POS category based on its context of use.

\subsubsection{Finnish UD}
The Finnish UD categories work with some mainly modal verbs (\textit{voida} `can', \textit{pitää} `should, have to', \textit{saattaa} `may, probably', \textit{täytyä} `have to', \textit{joutua} `get (into), have to', \textit{aikoa} `intend, aim', \textit{taitaa} `be able to, may', \textit{tarvita} `need', \textit{mahtaa} `might'; (in the negative) `can´t') as auxialiary verbs. However, number of Finnish modal verbs is much longer than this. In the list there is also a negative verb \textit{ei}, which can possibly affect the results, because the negative verb is separated from the main verb. The verb \textit{olla} `to be' is at the same time listed as an auxiliary and as the only copula verb.

The Finnish UD annotation does not use the term particle (PART) as one type of POS. It divides such words in the categories of adverbs, adpositions, conjunctions (coordinating and subordinating) and interjections.

Although some expressions behave similarly to articles in colloquial Finnish, such as \textit{se} `it', \textit{tä(m)ä} `this' and less frequent \textit{yks(i)} `one' (Hakulinen et al. 2004, § 1418), Finnish has no true articles. For this reason, DET does not appear as a separate POS in the Finnish version of UD.\footnote{See also Turku Dependancy Treebank \url{https://github.com/UniversalDependencies/UD_Finnish-TDT} and Finnish version of InterCorp \url{https://wiki.korpus.cz/doku.php/en:cnk:intercorp:verze16ud.}}

\chartQ{FI}{Finnish}{\legendnoPARTDET}

\subsubsection{Analysis}
The visualization of the two sets of Finnish data differs quite a lot. The reason for such a big difference could be the placement of the verb \textit{olla} `to be' and the negative verb \textit{ei} in the set A.  

The set A shows four clear tentacles of VERB, NOUN, ADJ and NUM, the ADJ tentacle being visibly shorter and close to NOUN one as these two categories are blending the most. Some participle forms tagged as VERBs can be not surprisingly found in the NOUN (e.g. \textit{lukittu} `locked --- nominative sg.', \textit{sisältävien} `containing --- genitive pl.') and ADJ (e.g. \textit{kiellettyjä} `forbidden --- partitive pl', \textit{liittyvistä} `joined --- elative pl.') tentacles. However, some other participles are tagged as ADJs. VERBs and NOUNs blend sometimes also due to homonymy: e.g. \textit{istuin} can be a verb form `I sat' as well as a noun `seat'. AUXs are part of the VERB tentacle.

Some of the clear adjectives were tagged as NOUNs, but they quite rightly appear within the ADJs tentacle (e.g. \textit{nätti} `nice --- nominative sg.', \textit{hidasta} `slow --- partitive sg.'). 
Ordinal numerals can be found from the ADJs tentacle, while the thin and clearly separated NUM tentacle consists mainly of the cardinals. ADVs do not form a tentacle, they form a concentrated group mainly near to VERB and NOUN tentacles, next to them there are also ADPs, which are more blending with NOUNs, as they are usually lexicalized forms of nouns (cf. \textit{väli} `space' and \textit{välillä} `in between').
PRONs are partly blending with ADJs (e.g. \textit{toinen} `other --- in different case forms', \textit{molemmille} `both --- allative pl'), partly with NOUNs (e.g. \textit{kaikkia} `all --- partitive pl', \textit{toisiamme} `each other --- partitive pl + possessive suffix 1pl'). 

In the set B the VERB, NOUN and ADJ tentacles have different shapes, they seem to be denser, shorter and closer to each other. The NUM tentacle is very thin and almost separated from the rest. Quite surprisingly, many TTAVA-participle forms are tagged here as ADJ, not VERB (e.g. \textit{naurettava} `laughable --- nominative sg', \textit{huomattavia} `noticeable --- partitive pl'), which can be a mark of the level of their lexicalization.   
In Finnish there are so few conjunctions, that there is no surprise not to see almost any of CCONJs and SCONJs in the data.

\subsubsection{Finnish conclusion}
Not surprisingly, NOUN and VERB tentacles are the prominent ones, followed by a shorter tentacle of ADJ. NUMs form a separate tentacle, but it is represented by cardinal numerals (mainly numbers/digits), while ordinals are as expected closer to ADJs. ADVs do not form a clear tentacle of their own and they are spread all over the categories, closest being VERBs and NOUNs. The same spread applies for PRONs and ADPs, but in their case, it seems to be the question of limited number of words in both categories.

\subsection{Russian}

\subsubsection{Russian POS}
\label{sec:russianPOS}
Parts of speech in the Russian language can be classified in various ways depending on the chosen criterion. Unlike the lexicological approach (which classifies parts of speech based on their ability to denote elements of extralinguistic reality, thus distinguishing words with nominative, demonstrative, auxiliary functions, and separately words expressing experiences), academic grammar and consequently school grammars and university manuals (e.g., \cite{svedova1980russkaja}, § 1111--1120)
 derive their classification from the grammatical structure of the language. They differentiate, based on shared grammatical features, between the following categories: fundamental (nouns, adjectives, numerals, pronouns, verbs, adverbs), auxiliary (prepositions, conjunctions, particles), and interjections. A word grammatically belonging to one POS can frequently appear in the function of another (e.g., an adjective functioning as a noun --- \textit{stolovaja} `canteen', a noun functioning as an adverb --- \textit{vek} `never'). A special category straddling POS is that of so called predicatives (cf. \cite{scerba1928casti}), which function as predicates and are sometimes classified among interjections, although some grammatical manuals define them as a separate part of speech expressing a state. E.g. \textit{Segodnya na ulitse \textbf{teplo}}. `It's warm outside today.' lit.: today on the street \textbf{warmly}.

\subsubsection{Russian UD}

The classification of parts of speech in UD differs significantly from traditional approaches in the following aspects: conjunctions are divided into two separate categories: subordinating (SCONJ) and coordinating (CCONJ). Moreover, unlike traditional Russian grammar, UD defines determiner as a separate category. Since Russian lacks articles, its grammar does not use the term determiners; most determiners are traditionally referred to as pronouns. In UD, attributive pronouns are labeled DET, whereas substantive pronouns are labeled PRON. UD also defines the category AUX (auxiliary VERB). In Russian, the only representative forms are those of the VERB \textit{byt'} `to be', which, unlike in other Slavic languages such as Czech, are never realized in the present tense. Compare the example of predicative above (\ref{sec:russianPOS}) where no verb is present and the sentence is interpreted as  being in the present tense (see english translation). In contrast, in the past and future tenses, the verb \textit{byt'} `to be' is expressed, e.g. \textit{na ulitse \textbf{bylo} teplo}  `It was warm outside today.' lit.: on the street \textbf{was} warmly.

\chartQ{RU}{Russian}{\legend}

\subsubsection{Analysis}

The visualization of Russian data shows clear tentacles for VERBs, NOUNs, ADJs, and NUMs. ADVs typically exhibit the least separation, with those expressing direction or spatial location (e.g., \textit{naruzhu} `outside', \textit{vverkh} `upward') being the most distinct from other POS. As for VERBs, participles are typically more isolated than infinitives (\textit{zhenit'sya} `to marry (of man)'); VERBs significantly shifted towards ADJs and NOUNs represent deverbative adjectives (\textit{okruzhayushchiy} `surrounding'), some of which can also function as NOUNs (\textit{pogibshikh} `the deceased'). The NUM tentacle is very isolated. Only the forms of the numeral \textit{odin} `one' are notably shifted towards adjectives. This is not surprising, as this numeral, unlike others, agrees grammatically with the counted object. Russian commonly uses both long-form and short-form adjectives. 

Although traditional grammars and UD classify them under a single ADJ category, the visualization clearly reveals distinct characteristics of these groups. Short-form adjectives in Russian, unlike their long-form counterparts, cannot function as attributive modifiers (compare \textit{molodoy muzhchina} / \textit{*molod muzhchina} `young man'), but are typically used as nominal predicates (\textit{muzhchina molod} `the man is young' lit: man young). The visualization indicates that the overlap between VERBs and ADJs exclusively involves adjectives in their short form (\textit{nuzhen} `needed', \textit{soglasna} `agrees'). A significant overlap of NOUNs and ADJs is observed with words that possess adjectival morphological properties but have become established as nouns (e.g., \textit{chayevyye} `tips'). The overlap between VERBs and NOUNs is minimal, with infinitives showing the most tendency towards NOUNs. Predicatives (e.g., \textit{vidno} `visible' and \textit{slyshno} `audible') are shifted towards VERBs, but interestingly not more so than typical full-meaning ADVs. Generally, there are not many predicatives present in the data.

\subsubsection{Russian Conclusion}

It is entirely expected that Russian, as a fusional language, exhibits very distinct tentacles of NOUNs, ADJs, and NUMs, with ADVs being notably less distinct. The overlaps among various POS also correspond to the morphosyntactic characteristics of Russian. The proximity of short-form ADJs and VERBs correlates with the absence of an auxiliary verb in the present tense. Expected overlaps also occur due to shifts between POS, typically involving the substantivization of ADJs. The only surprising aspect might be the position of predicatives, which, although on the periphery of an otherwise compact cluster of ADVs, might have been expected to show a more pronounced shift towards VERBs.

\subsection{English}

\subsubsection{English POS}
English grammars generally define eight parts of speech (sometimes also called word classes): nouns, determiners, adjectives, verbs, prepositions, adverbs, conjunctions and interjections (\cite{cambridgegrammar}; \cite{aarts-2011}). The number of word classes can vary based on the definitions of a word class and word class subsets. For example, in (\cite{cambridgegrammar}) that does define eight parts of speech, pronouns are defined as a subset of nouns, while certain grammars define them as a separate part of speech (e.g. \cite{quirk}, which defines ten parts of speech). Some grammars also define main and secondary word classes, with primary being nouns, verbs, adverbs and adjectives.
English definitions of POS are based on syntax, semantics and partly morphology (some grammars give a list of suffixes that are usually used in certain word classes, such as \cite{cambridgegrammar}). 

\subsubsection{English UD}
English UD classification varies in some ways. Firstly, UD defines auxiliary verbs as their own separate word class. These auxiliary verbs significantly differ from those defined by the grammars, including not only the usual verbs \textit{be, have} and \textit{do}, along with modal auxiliaries \textit{can, may, shall, will} and \textit{must}, but also the passive \textit{get} and other modal verbs such as \textit{should} or \textit{dare}. Pronouns and numerals (which are either defined as nouns or determiners) are also defined as their own POS. Conjunctions are separated into coordinating and subordinating. UD also defines particles. These include possessive markers (e.g. \textit{`s}), predicate negations (e.g. \textit{not, nt}) and the infinitive marker \textit{to}.\footnote{The authors themselves are considering reworking this specific category.} Lastly, UD also uses adpositions, which include specific uses of the words \textit{in} and \textit{to}.

\chartQ{EN}{English}{\legend}

\subsubsection{Analysis}
The English data visualisations show, unsurprisingly, three main tentacles of NOUNs, VERBs and ADJs. They also define smaller tentacles of NUMs and ADVs.
Blending/transition of tentacles: Compared to other languages analysed in this study, the English visualisations show the three main tentacles in a more tight position, with a lot of blending especially between the VERB and NOUN tentacle. The words that cause this blending are mainly the polyfunctional ones: the majority of them either being verbal nouns with the suffix -ing, which also serve as the present continuous tense form (e.g. \textit{clanking}, \textit{drowning}, \textit{mixing}), infinitive forms of verbs that can also be used as nouns (e.g. \textit{laugh}, \textit{quote}, \textit{care}), or verbs in the 3rd singular person of present indicative form which can also be used as plurals (e.g. \textit{attempts}, \textit{exhales}, \textit{kisses}). Some blending also occurs between the NOUN and ADJ tentacles. The words that blend the most are mostly polyfunctional (e.g. \textit{secret},\textit{ commercial}, \textit{alternative}) or ones with typical collocations or monocollocability (e.g. \textit{vocational}, which collocates mainly with the words training or education).  Slight blending between the ADJ and VERB tentacles occurs as well, mainly comprised of verbs with the -ed suffix which also act as adjectives (e.g. \textit{integrated},\textit{ related}, \textit{impressed}) and, again, polyfunctional words (e.g. \textit{elaborate}, \textit{idle}, \textit{welcome}). Even though ADVs form their own tentacle, facing away from the others, some ADVs are not part of the tentacle and blend either into the cluster of other POS that do not have their own defined space, or into the other tentacles. Most of the words that blend elsewhere are spatial ADVs (e.g. \textit{abroad}, \textit{nearby}, \textit{north}), but also those with monocollocability or typical collocations (e.g. \textit{ago} is blended with NUMs).

\subsubsection{English conclusion}
Overall, the visualisation shows clear tentacles of the four main POS (NOUN, ADJ, VERB, ADJ), which is not surprising. It is curious that the ADJ, NOUN and VERB (and especially the last two) tentacles are so close to each other. Even though they do have their individual syntactic functions, it appears that semantically, they might be closer to each other than the definitions in the English grammars might imply. Of course, the category of the prototypical NOUNs that are well separated from the VERBs is irrefutable, but the definition of those that do blend could be controversial. Moreover, NUMs having their own separate tentacle is surprising, not having their own POS, but also in line with what we're seeing in other languages analysed.

\section{Conclusions}

This study has presented a visualization that helps navigate the semantic latent space of parts of speech as understood by Universal Dependencies. The visualization is based on corpus data and works with embeddings that are further reduced using a POS classifier based on neural networks.

It should be noted that the POS classifier itself is not the aim of this study. The classifier performs reasonably well (see confusion matrices in the Supporting material on OSF; the visual mapping could not function without a working classifier), but if we wanted to create the best-performing classifier possible, the topology of the neural network would be different. 
Rather, the main output is the visualization itself. 

These results were further analyzed, though the analyses presented were not intended to be a comprehensive description, but rather a guide for reading the data and an encouragement for readers to discover their own meaning within them.

We have chosen languages that are all from the European area and can be considered culturally close, though not always linguistically related: four languages are from the Indo-European family (one Romance, one Germanic, and two Slavic) and one Uralic. This allowed us to examine how the traditions differ between two closely related languages (Czech and Russian) or how tradition differs in a language that is non-Indo-European, with a linguistic tradition slightly adapted to this fact. While Finnish is by some authors believed to lack clear definitions of parts of speech, the results did not appear more chaotic than those of the other languages.

It would be interesting to conduct systematic cross-linguistic comparisons and add more languages, particularly non-European ones, but we leave this to our readers, as we have neither the expertise nor the resources for such an undertaking.

This method is applicable to other frameworks than Universal Dependencies and can thus serve to compare different traditions not only cross-linguistically but also within a single language.

\section*{Data Availability}
All data, scripts, and visualizations, along with further analyses of the models (F1 scores, confusion matrices, etc.), are available at \url{https://osf.io/gm6zd/}. The interactive charts are also mirrored at \url{https://hypertext.cz/pos}.

\section*{Acknowledgements}
The work has been supported in part by the Ministry of Education, Youth and Sports of the Czech Republic, Project No. LM2023062 LINDAT/CLARIAH-CZ. 

The implementation of the Corpora in the KonText was supported by the Czech National Corpus project (LM2023044) funded by the Ministry of Education, Youth and Sports of the Czech Republic within the framework of Large Research, Development and Innovation Infrastructures.

\printbibliography

\end{document}